# A VIEW-DEPENDENT ADAPTIVE MATCHED FILTER FOR LADAR-BASED VEHICLE TRACKING

Daniel D. Morris, Regis Hoffman, Paul Haley
General Dynamics Robotic Systems, (GDRS),
1501 Ardmore Blvd, Pittsburgh, PA 15221, USA
{first initial last name}@gdrs.com

**ABSTRACT**
LADARs mounted on mobile platforms produce a wealth of precise range data on the surrounding objects and vehicles. The challenge we address is to infer from these raw LADAR data the location and orientation of nearby vehicles. We propose a novel *view-dependent adaptive matched filter* for obtaining fast and precise measurements of target vehicle pose. We derive an analytic expression for the matching function which we optimize to obtain target pose and size. Our algorithm is fast, robust and simple to implement compared to other methods. When used as the measurement component of a tracker on an autonomous ground vehicle, we are able to track in excess of 50 targets at 10 Hz. Once targets are aligned using our matched filter, we use a support vector-based discriminator to distinguish vehicles from other objects. This tracker provides a key sensing component for our autonomous ground vehicles which have accumulated hundreds of miles of on-road and off-road autonomous driving.

## 1. Introduction

Vehicle detection, tracking and trajectory estimation are key problems that must be addressed to avoid collisions and achieve safe autonomous navigation. For this we leverage LADAR sensors mounted on a mobile platform which produce precise, real-time, scanned range estimates of the surroundings. However, these data do not come labeled with target type nor position on the target, and so individually the range estimates provide very weak constraints on target pose. To use the range measurements in a kinematic vehicle tracker, they must be upgraded into measurements of vehicle pose. This paper proposes a novel *view-dependent adaptive matched filter* (VDAMF) that can be optimized over range measurements to obtain vehicle pose. Other techniques have been proposed for this measurement task, but the VDAMF combines advantages in robustness, simplicity and computation efficiency.

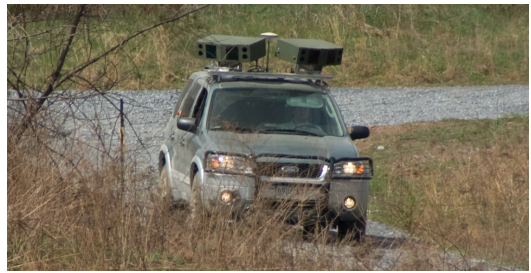

Fig. 1 One of our autonomous vehicles with LADAR sensors on the roof.

While there is a relatively large body of work on image-based vehicle detection and tracking, (see [1] for a summary), there has been much less analysis of how LADAR data can best be used for this task. Scanning LADARs have traditionally been used for mapping and analyzing stationary objects, (see [2], [3], [4], [5], [6], and [7]). Recently, there have been a number of extensions to detect moving targets using techniques such as global segmentation, [8], cluster similarity, [9], feature detection, [10], [11], [12], [13], [14], [15], [16], and model fitting, [17]. Of the techniques that explicitly seek to detect vehicles, most rely on finding straight-edge features in the data and infer vehicle positions from one or more of these edges (see [11], [12], [13], [14] and [15]). We previously implemented an "L-shape" fitting technique to find vehicle pose, [12], and found that with noisy and cluttered data there are many ambiguities in fitting vehicle edges. Robust fitting entails enforcing a variety of geometric and visibility constraints as well as using heuristics and *ad hoc* rules to keep the number of edge-fits manageable. Strategies, like requiring a minimum edge length, preferentially fitting points close to the ground so as not to detect vehicle cabs, enforcing visibility constraints to avoid viewing a vehicle corner from the inside, and others, are all useful; but they add significantly to the complexity of RANSAC or other robust procedure used to fit edges.

Another approach, described in [17], uses rectangular vehicle models and simultaneously estimates pose, speed, and shape parameters in a Bayesian filter. There is some similarity to ours in that LADAR hits are integrated over a cost that depends on target position and size. It has some advantages over our approach in that it explicitly models occlusion and does not require pre-clustering of points. This approach seeks to do detection, filtering and tracking in a single optimization, and as a consequence, the motion model is very simple (to keep the total number of parameters manageable), and optimization requires statistical sampling.

In contrast to [17], we focus on the measurement task, separating this from the tracker which can use a more sophisticated variable-axis kinematic model, [12], enabling better tracking of vehicles through curved trajectories. Also our measurement model takes into account view-dependent self-occlusion effects, and our cost function is analytic, enabling fast optimization and perturbation-based uncertainty modeling. While we require pre-clustering of LADAR points, this can be an over-segmentation as our model merges target clusters. In contrast to edge-fitting methods with many fitting rules, our measurement is an efficient gradient-descent-based optimization. The complexity is up-front in the VDAMF definition, and this filter is fairly simple.

A summary of our paper is as follows. In Section 2 we describe the tracking architecture followed by data modeling in Section 3. In Section 4 we motivate and propose our VDAMF. Next we derive an analytic solution for the matching function and address its optimization. We describe how to handle partial visibility loss by decoupling the target center estimates from target dimension estimates. We present our method for discriminating vehicles from clutter objects in Section 5. Finally we present some sample results, discuss limitations and conclude.

## 2. Tracking Architecture

The context of this work is the need for autonomous ground vehicles, as in Fig. 1, to navigate safely in cluttered environments. To achieve this, other objects in the local vicinity, and in particular other vehicles, need to be detected and tracked and their future positions predicted. A flow chart of our tracker that achieves this is shown in Fig. 2. We briefly summarize the components here.

Tracking starts with a frame of LADAR data sampling the field of view and transformed into fixed world coordinates using a precise, onboard inertial navigation system enabling operation independent of sensor platform motion. The pre-processing for each frame involves removing ground points and segmenting the remaining points into clusters, one of which is shown in Fig. 3(*a*). Our main requirement is that this be an over-segmentation such that clusters lie in at most one target. In this regard we assume vehicles are separated from other objects by at least 1 meter.

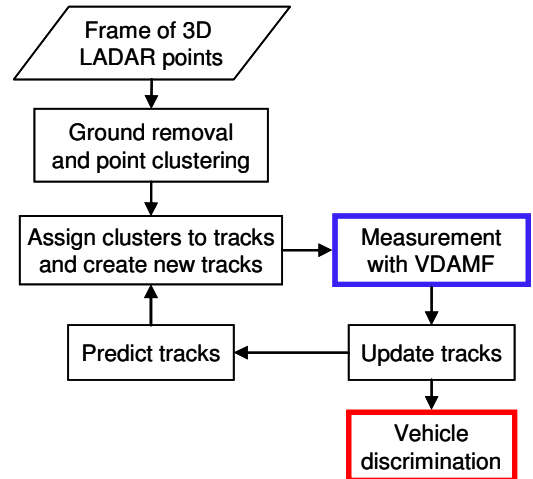

Fig. 2. Architecture of our vehicle tracker. This paper focuses on the measurement step, with a short description of vehicle discrimination.

Using predicted track poses and sizes, clusters are assigned to tracks using a simple auction scheme. Multiple clusters can be assigned a single target if they are consistent with its projected position. Those not claimed by current tracks are used to start new tracks.

Next target poses are measured using the adaptive matched filter proposed in this paper. Since *a priori* it is unknown which objects in the field of view are vehicles, we track all objects and leave vehicle discrimination to a post-tracking stage. The VDAMF is flexible enough to track most non-vehicle objects in addition to vehicles. We restrict analysis to horizontal planar motion.

Since the measurement step is self-contained, we are free to select from a variety of trackers for the update and prediction steps. Our choice is an Extended Kalman Filter that uses a Variable-Axis Steering Model described in [12]. Once objects are tracked our final step is to determine whether or not they are vehicles. For this we use a support vector-based discriminator described in Section 5.

## 3. Data Modeling

Our tracker must work on a variety of LADARs with different scan patterns, and so we make minimal requirements on the data. We assume data are acquired by a 2D scanning LADAR[1] and each frame provides a roughly uniform sampling of the field of view. The key question we address is how to infer from these points the vehicle pose.

The measured LADAR points are where the laser rays intersect surfaces on the object. The exact 3D point depends on physical surface properties such as reflectivity, the beam width, as well as algorithmic processes within the LADAR. We do not wish to model objects at such a fine scale and instead make the following simplification. Given a LADAR hit, we model the probability density of the local target surface around this point with a Gaussian centered at the LADAR point. The variance of this Gaussian includes both the uncertainty in range inherent in the LADAR, and also the variability of the target surface which may be significantly greater. Our target signal is thus a standard 2D Gaussian mixture model, representing the sampled surface of a vehicle, as illustrated in Fig. 3(*b*).

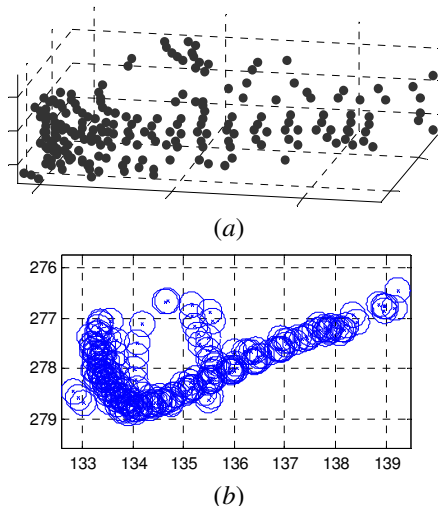

Fig. 3 (*a*) A cluster of 3D LADAR points belonging to a target vehicle from a single frame. (*b*) Our representation is a Gaussian mixture model of points projecting into the horizontal plane.

## 4. VDAMF

Matched filters are commonly used in signal processing, image processing, and sonar and radar trackers. A target is found in a noisy signal by designing a filter that matches the target signal signature and convolving this over a noisy signal. This maximizes the signal to noise ratio enabling the target to be identified from the background clutter and its position estimated. We restrict the purpose of the VDAMF to the second of these: determining target position, orientation and size. This provides alignment for a later 3D discriminator, described in Section 5, that separates vehicles from clutter.

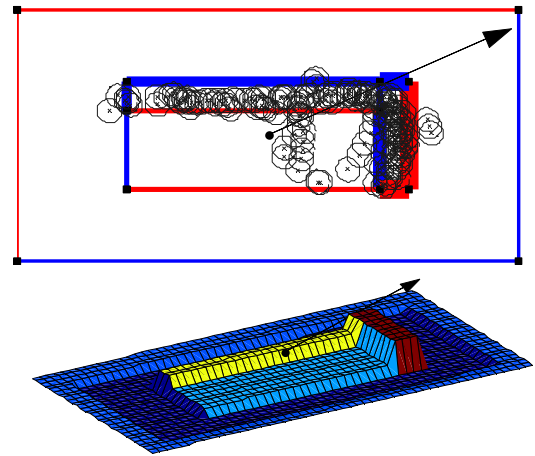

Fig. 4. (*a*) LADAR hits on a vehicle (crosses) lead to a Gaussian mixture model for the visible vehicle surfaces (black circles). To approximate this density, we create a 2D matched filter by summing 4 rectangular regions. On the outside is a negative rectangle. Inside this, a positive rectangle covers the vehicle interior. The two visible sides are represented with rectangles having magnitudes dependent on the ray to the sensor shown with an arrow. Red lines indicate negative steps from the top left, and blue lines positive steps, in the density function. (*b*) The same filter shown as a surface plot.

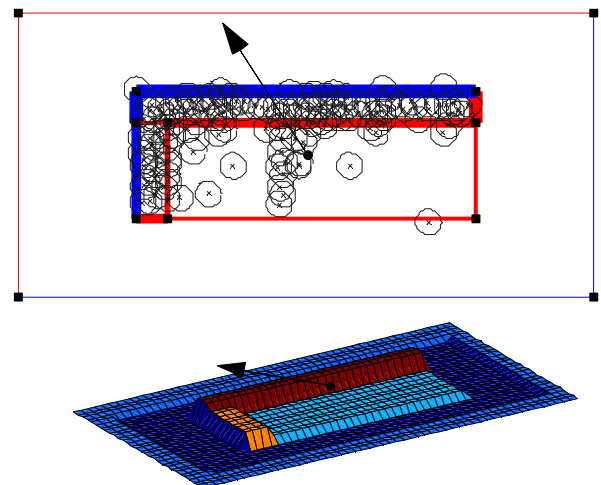

Fig. 5. Another view of the vehicle same shown in Fig. 4. LADAR hits cluster mostly along the side and rear of the vehicle. The optimized matched filter is shown overlaid, as well as in a surface plot below.

The following are some key challenges we face with LADAR data and which we design the VDAMF to address. The first is that vehicle size and shape vary greatly between vehicle types. The filter must be general enough to match most vehicle signatures, and at the same time minimize loss of precision due to lack of specificity. We address this with a simple and general filter whose size is adaptively estimated during the optimization process; hence "Adaptive" in the name. Next, the number of hits on a target, and their resolution, varies with the square of the range. This is naturally handled by normalizing our filter. Finally a large portion of the vehicle (at least half) is self-occluded at any given time and surface sampling will vary depending on incident angles. To account for these self-occlusions and sampling variations, we make our filter response a function of the relative viewing angle; hence "View-Dependent".

---

[1] In fact the proposed filter will also work with a horizontal line-scanning LADAR, although with reduced robustness.

An ideal matched filter will exactly equal the target signal. We seek a matched filter that will best approximate the signal, an example of which is shown in Fig. 4(*b*). We assume that vehicles typically have a rectangular shape (viewed from above). The number of hits on any non-occluded surface region is proportional to the solid angle it subtends at the sensor. This means the majority of hits will be on the vehicle sides facing the sensor and will be proportional to the cosine of the angle between the surface normal and the sensor. We expect additional LADAR hits in the interior region of the rectangle, although their distribution will depend on the particular vehicle type, and so for simplicity we assume a uniform distribution in the interior. By projecting points into the horizontal plane we achieve invariance to vehicle height. Finally, we add a negative boundary around our vehicle. Ideally there will be no points in this region and so it will not affect position estimation. Rather, it is needed for vehicle size estimation, as it ensures a tight boundary around the LADAR points. This leads to our proposed adaptive matched filter, illustrated in Fig. 4 and Fig. 5, which combines an interior region, two visible edges, and a negative external boundary.

Table 1 Specification of our adaptive matched filter. It's size depends on three parameters: vehicle length *l*, width *w*, and sensor angle $\beta$.

| Region | Dimensions | Weight |
|---|---|---|
| Surround | $(l+1.5m) \times (w+1.0m)$ | -0.25 |
| Interior | $l \times w$ | 0.1+0.25 |
| Side edge | $l \times 0.6m$ or $(l-0.8) \times 0.6m$ | $\sin(\beta)$ |
| End edge | $w \times 0.8m$ or $(w-0.6) \times 0.8m$ | $\cos(\beta)$ |

Our filter is a combination of four uniform-height rectangular regions. We use rectangular regions both because they well approximate vehicle components, and because of their convenient analytic properties. The dimensions and weights of each region are shown in Table 1. The key parameters that its shape depends on are vehicle length *l*, width *w*, and sensor angle $\beta$. The latter is the difference between the orientation angle of the vehicle and the angle to the sensor. When the target vehicle is pointing directly towards the sensor, this is zero. The weights are chosen by hand to be representative of the LADAR point distribution we observed, but in the future they could be learned. Their absolute values do not matter as the filter is normalized. Only the two visible edges, (whose weights are greater than zero), are included. The weight-dependence on sensor angle models visibility and self-occlusion effects. The edge whose weight is greatest spans the whole length or width, as illustrated in Fig. 4 and Fig. 5. When integrated over the target density, Eq.(4), the results is a continuous function that is easily optimized.

### 4.1 Filter Matching Definition

In filter matching we seek to optimize a 5-dimensional space: pose which we refer to as: $t = (t_x, t_y, t_\theta)$, and vehicle size which we indicate by: $\mathbf{w} = (w, l)$ containing width, *w*, and length, *l*. Performing discrete convolutions over a 5-dimensional space is not practical. Even performing convolutions over just the 3-dimensional pose space will be slow as the filter response is very sensitive to orientation, requiring building filters for roughly 1-degree increments. It is for this reason that we derive an analytic cost to the filter matching process that can be easily optimized. In addition, an analytic cost enables us also to derive a perturbation-based uncertainty measure for our matched solution which is very useful for Kalman Filtering.

Given a noisy signal, $p(x)$, and a filter, $s(x)$, equal to the ideal signal, then a 1D matched filter finds the offset, *t,* given by:

$$\arg\max_t \int p(x)\tilde{s}(x-t)\,dx. \quad (1)$$

Here we have used the normalized filter $\tilde{s} = \alpha s$, where $\alpha$ is the normalizing constant. Normalization is important when optimizing over parameters that affect filter response, such as filter size.

We generalize the 1D filter matching process in Eq. (1) as follows. Our signal is defined over the plane and consists of a sum of Gaussians:

$$P(x, y) = \sum_i N(\mathbf{m}_i, \sigma; x, y) \quad (2)$$

where *N* indicates a normal distribution with mean $\mathbf{m}_i = (x_{mi}, y_{mi})$ and standard deviation $\sigma$. Our filter is also defined over the plane and is a function of pose and size parameters: $s(t, \mathbf{w}; x, y)$. The normalization factor

is:

$$\alpha = 1 \Big/ \sqrt{\iint s^2(t,w;x,y)dxdy} \quad (3)$$

Then the matched filter response to the signal given parameter values $(t,w)$ is obtained as the integral over the plane:

$$M(P,s;t,w) = \iint P(x,y)\tilde{s}(t,w;x,y)dxdy \quad (4)$$

### 4.2 Analytic Expression of the Matching Equation

Vehicle position, pose and size estimation is achieved by maximizing the matched filter response, Eq. (4), as a function of $(t,w)$. To perform this efficiently, in this section we derive an analytic cost. This analytic expression has added benefit in that it makes perturbation analysis straightforward allowing us to derive a covariance estimate for our estimate.

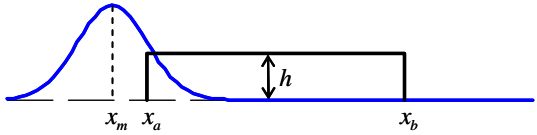

Fig. 6. An example of a Gaussian signal and a one-dimensional box filter.

The difficulty for analytic optimization is that each point must be integrated over the target. To achieve this, we use a trick similar to integral images. We will derive it for 1D signals and extend it to 2D.

Let the signal be a 1D Gaussian, $N(x_m,\sigma)$, with mean, $x_m$, and standard deviation $\sigma$. The integral of this over the unit step function occurring at $x_a$ is given analytically by:

$$I_{ma}(x_a,x_m) = \int_{xa}^{\infty} N(x_m,\sigma)dx = \frac{1}{2}\left(1 + \mathrm{erf}\left(\frac{x_m - x_a}{\sqrt{2}\sigma}\right)\right) \quad (5)$$

The integral of this signal over the box function in Fig. 6 is obtained by adding integrals over a positive and negative step functions, weighted by the filter height $h$, namely:

$$I_{box} = h(I_{ma} - I_{mb}) \quad (6)$$

Now extending to 2D, consider the problem of integrating a Gaussian over a rectangular region illustrated in Fig. 7. Let $I_d$ be the integral of the Gaussian over a unit-height quarter-plane indicated by the shaded region, and $I_a, I_b, I_c$ be the integrals over the remaining quarter planes extending to positive infinity. The integral over each quarter plane is the product of $x$ and $y$ 1D integrals:

$$I_d = I_{dx}I_{dy}, \quad (7)$$

where $I_{dx}$ and $I_{dy}$ are given by Eq. (5). Then the integral over the rectangle of height $h$ is:

$$I_{rec\tan gle} = h(I_a + I_d - I_b - I_c). \quad (8)$$

The consequence of Eq. (8) is that so long as the matched filter is constructed from rectangular regions, its response to the data described in Eq. (4) can be expressed analytically without the need for discrete convolutions. We note that rectangle edges must be parallel to the axes. Hence, instead of rotating the filter to match the data, we transform the data to match the filter. That is, LADAR point centers $m_i$ are translated with respect to chosen coordinates $m_0$ (typically the expected vehicle location) and rotated with the negative filter angle $t_\theta$:

$$^0m_i = R(-t_\theta)(m_i - m_0) \quad (9)$$

where $R(-t_\theta)$ is a $2\times 2$ rotation matrix.

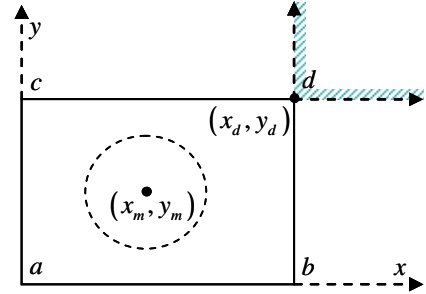

Fig. 7. A Gaussian point being integrated over a uniform rectangular region *abcd*. The result is the sum of functions calculated at the 4 corners, similar to integral images.

The integral over the filter is the corresponding linear combination of integrals calculated at each corner. Hence Eq. (4) becomes:

$$M(t,w) = \alpha \sum_{i,j} c_j I(s_j, {}^0m_i) \quad (10)$$

where the sums are over all points, $i$, and filter corners, $j$, and each integral term $I$ is the product of two 1D integrals from Eq. (5). In order to optimize $M$, it is important to explicitly account for the effect of each parameter on the right hand terms of Eq. (10), and so their dependency is specified as follows. As seen in Eq. (9), the point centers $^0m_i$ depend on $t_\theta$. The coordinates of each corner point $s_j$ depends on the parameters $(t_x,t_y,w,l)$. The coefficients $c_j$ determining the weight of each rectangular region, (see column 3 of Table 1). These depend on $\beta = (t_\theta - \varphi)$, the difference between the target orientation and the viewing angle $\varphi$. Finally the normalization coefficient depends on the full parameter set $(t,w)$, as well as the viewing angle $\varphi$, which we approximate as a constant during optimization.

### 4.3 Filter Optimization

The measurement process involves maximizing the matched filter in Eq. (10) as a function of its five parameters: pose and size. It is not feasible, nor necessary, to exhaustively search this space at each LADAR scan. Rather LADAR hits that are above the ground surface are initially clustered and the boundaries of each cluster provide a starting state for filter optimization.

Now our analytic expression, Eq. (10), is readily differentiated to first and second order enabling us to directly calculate the gradient, $\nabla_t M$, and Hessian, $\nabla_t^2 M$. We used these, and a Levenberg Marquardt procedure, to optimize the filter. We note that to obtain quadratic convergence, it is necessary to include all the second order terms in the Hessian, (rather than use just a first order approximation for it). In addition, the Hessian becomes non-positive semi-definite when distant from the optimum, and so it is important to ensure that the diagonal term added to it in the Levenberg Marquardt procedure is large enough to make it positive definite. Given these adjustments, we found fast convergence in roughly 4 to 8 iterations.

### 4.4 Perturbation Analysis

The result of maximizing Eq. (10) is an estimate $\hat{t}$ for the vehicle position and orientation. To be useful in a tracking filter, we also would like a covariance estimate of $\hat{t}$. We obtain this through perturbation analysis. At the minimum we have

$$\nabla_t^2 M \, \Delta t = -\nabla_t M \, , \quad (11)$$

which will be zero. Now if one of the LADAR hits, $p_i$, is perturbed in the plane by $\Delta p_i$, then to first order this will result in a perturbation $\Delta t$:

$$\Delta t = -\left(\nabla_t^2 M\right)^{-1} \nabla_{tp} M_i \Delta p_i . \quad (12)$$

Here $M_i$ is the component of Eq. (10) relating to point $p_i$, and we define a 3x2 matrix:

$$\nabla_{tp} M_i \equiv \sum_j \left( \frac{\partial \nabla_t M_i}{\partial x} \quad \frac{\partial \nabla_t M_i}{\partial y} \right) \quad (13)$$

with the columns being the $x$ and $y$ partials the gradient vector $\nabla_t M$ for point $p_i$. Modeling LADAR hit position uncertainties as independent with uniform variance, $\sigma_p^2$, we obtain an expression for the covariance of $\hat{t}$ by combining perturbations from all points:

$$R_t = \sigma_p^2 \left(\nabla_t^2 M\right)^{-1} \left( \sum_i \nabla_{tp} M_i \nabla_{tp}^T M_i \right) \left(\nabla_t^2 M\right)^{-1} \quad (14)$$

Notice that the magnitude of the matched filter cost, $M$, cancels implying that the covariance is independent of this.

### 4.5 Visibility Corrections

One complication with including size parameters in the estimate is that when the observed width or length changes, the measured vehicle center position also changes. A common situation for this is when a target vehicle rotates so that only the rear or front is visible; in these cases the measured length will be much smaller than the actual length, and unless this is handled properly it could create a phantom velocity of the target vehicle. In [15], this problem is addressed by using a fixed anchor point. However, this entails adding two additional parameters to the state vector; a significant increase in dimensionality. In contrast, our approach is to detect when visibility loss occurs and in those cases use one or more visible corners of the vehicle as fixed points. Partial visibility loss is detected by comparing the current measured length and width to a long-term robust average; when these do not match, a visibility loss on the length or width is detected.

When visibility loss is detected, we need to adjust either the state in the tracker or the measured position $\hat{t}$, so that their positions would match if there were no noise. We found that adjusting the state resulted in more stable tracks, and so using the corner closest to the sensor as a fixed point we offset the state translation components to compensate for change in measured target size. In addition, the measurement covariance, $R_t$, should be transformed, as this is calculated assuming full edge visibility. To transform this we introduce a change in variables. Our new variables include the same target orientation angle, $t_\theta$, and the remainder are defined in the local target coordinate system. They are the change in $x$ positions of the front and rear, $\Delta x_f$ and $\Delta x_r$, and the change in $y$ positions of the right and left sides, $\Delta y_r$ and $\Delta y_l$. The transformation from these to the center coordinates is given by:

$$\begin{pmatrix} \Delta t_x \\ \Delta t_y \end{pmatrix} = R(t_\theta) \begin{pmatrix} \frac{1}{2} & \frac{1}{2} & & \\ & & \frac{1}{2} & \frac{1}{2} \end{pmatrix} \begin{pmatrix} \Delta x_f \\ \Delta x_r \\ \Delta y_r \\ \Delta y_l \end{pmatrix} \quad (15)$$

Where $R(t_\theta)$ is a 2D rotation matrix. The change in width and length are: $\Delta l = \Delta x_f - \Delta x_r$, and

$\Delta w = \Delta y_r - \Delta y_l$.

Working in this local coordinate system, it is straight forward to eliminate non-visible components of the measurement covariance. For example, viewing a vehicle head on will result in the front being well sampled with no or few hits on the sides and none on the rear. Fitting these points will give information on the front $\Delta x_f$, and the sides, $\Delta y_r$ and $\Delta y_l$, but not the rear $\Delta x_r$. Thus we eliminate the component corresponding to $\Delta x_r$ in the calculation of $\mathbf{R}_t$.

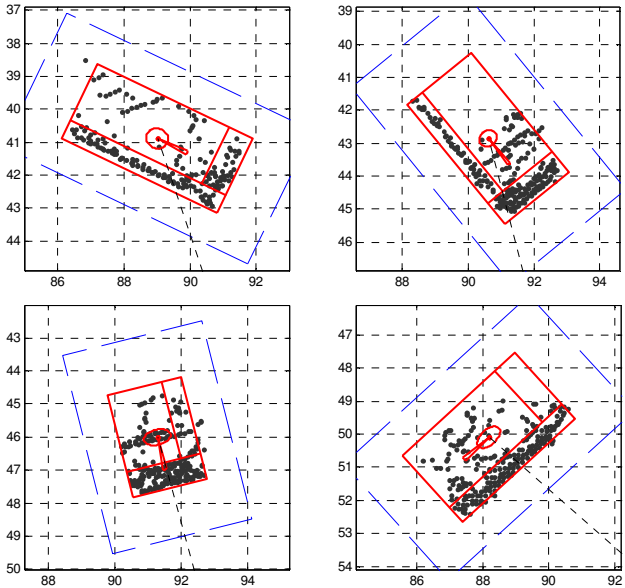

Fig. 8. Results of optimizing the matched filter for a sequence of frames taken as a target vehicle circles in front of the LADAR sensor. The black dashed line shows the direction to the sensor. The large red rectangle indicates the measured position, pose and size of the vehicle in each frame. The visible edge regions of the rectangle are indicated in each case, and the negative surround region is plotted as a blue-dashed rectangle. A 1-sigma ellipse around the center-point indicates the position uncertainty, and a 1-sigma pie segment indicates the angular uncertainty. Notice in each case the length and width adapts to the visible portion of the vehicle.

## 5. Vehicle Discrimination

We briefly summarize the shape-based vehicle discriminator originally proposed in [15]. This is the final step of our tracker, see Fig. 2. It is used to tell the predictor and path planner which objects are likely to be vehicles.

The concept is first to align all observed objects in a canonical coordinate system using the VDAMF. We select the closest corner to the sensor as the origin, the side of the filter as the $x$ axis and the end of the filter as the $y$ axis. We assume lateral symmetry for vehicles, and so as necessary mirror the points so that they lie along the positive $y$ axis. Vertically points are aligned relative to the local ground height. Once points are transformed into this coordinate system they are binned into a fixed-size 3D grid as illustrated in Fig. 11(*a*). A linear Support Vector Classifier is trained on both positive and negative examples of vehicles and the result is illustrated in Fig. 11(*b*). Finally the classifier is tested on ground-truthed data with performance shown in Fig. 11(*c*).

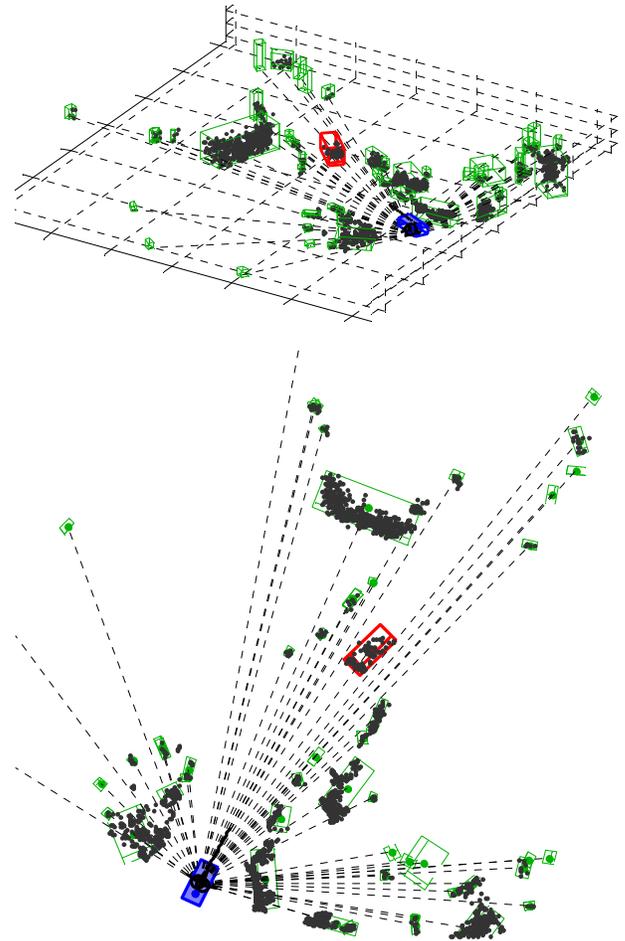

Fig. 9. A moving sensor platform (blue rectangle) following a target vehicle through wooded terrain. All objects are fit with the VDAMF and tracked. By aligning data to these pose estimates, the vehicle discriminator correctly identifies the target vehicle as a vehicle (red rectangle) and the rest of the objects as non-vehicles (green rectangles). Dashed lines are drawn from each object to the sensor. A 3D view (above) and a top-down view (below) are shown.

## 6. Results

The VDAMF was implemented and used as the measurement step of an Extended Kalman Filter for tracking vehicles and clutter. Computational requirements are low enabling us to track 50 targets at 10 Hz on a single core of a conduction-cooled Core 2 Duo. Some examples of measuring a vehicle's pose and size with the VDAMF are provided in this section. The covariance estimates are illustrated with ellipses in each case.

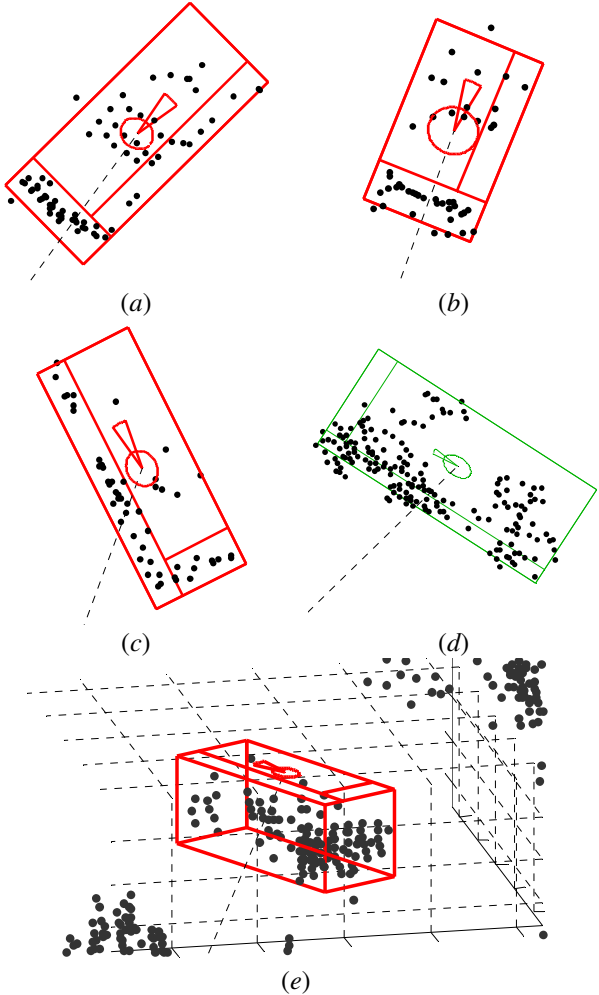

Fig. 10. (*a*) – (*c*) Close-up views of the VDAMF fitting the LADAR hits of the target vehicle in Fig. 9. (*d*) The filter fit to a non-vehicle object, in this case some bushes. (*e*) 3D illustration of the data in (*c*).

Fig. 8 shows snap-shots of the VDAMF fit to a Humvee driving in a circle. The sampled points on the vehicle depend strongly on the viewing angle and are fit well.

Fig. 9 shows an example of a pickup truck being followed by the sensor platform along an unpaved road through wooded terrain with regions of dense brush. Here the matched filter is used to estimate pose of all visible objects including the target vehicle and the bushes and trees. The fitting is agnostic to object type, although naturally matching score tends to be best for vehicle-shaped objects. Close-ups of a number of frames of the tracked vehicle and of tracked clutter from Fig. 9 are shown in Fig. 10. This illustrates robustness to noisy data and generality to a wide variety of objects. Our discriminator, Fig. 11, determines which targets are vehicles.

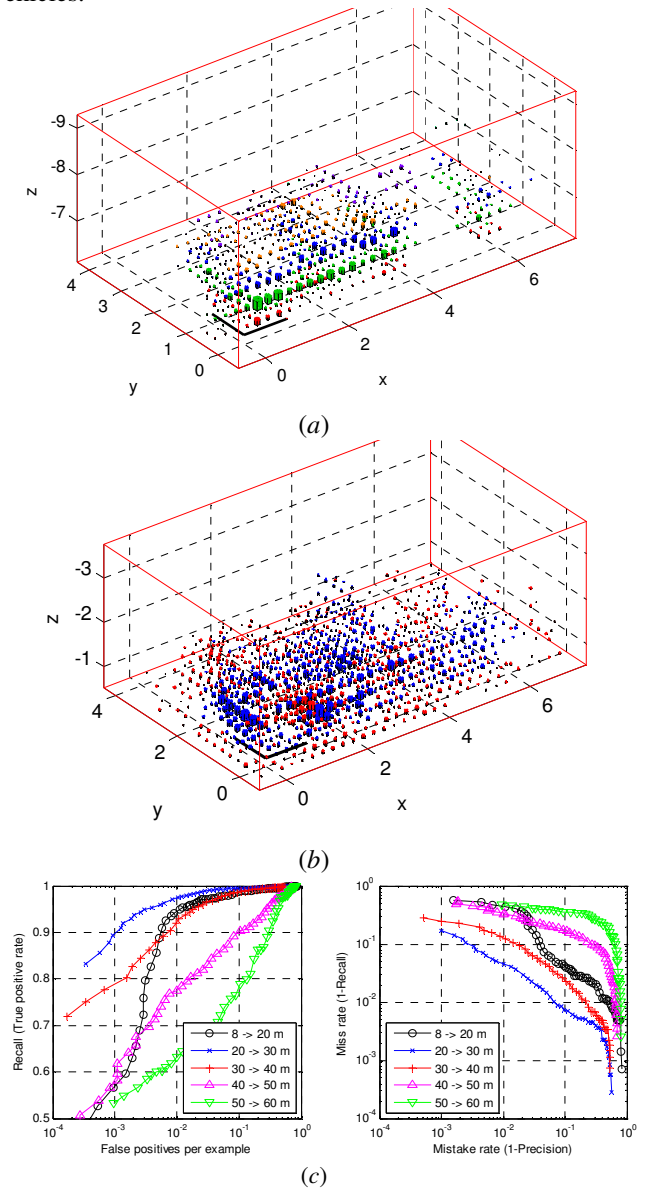

Fig. 11 (*a*) 3D hits on a vehicle in its local coordinate system obtained by the VDAMF are binned. (*b*) A linear Support Vector Classifier trained on 10,000 positive and 20,000 negative examples is illustrated here. Blue indicates positive values and red negative in the normal to the separating hyperplane. (*c*) ROC and Detection Error Tradeoff curves showing performance of the discriminator on vehicles and clutter.

## 7. Limitations

A number of trade-offs have been made in seeking robustness, generality and accuracy. Some viewpoints of some vehicles can have poor matches to our model, particularly at long range when there are few hits on target. The target vehicle is still fit, but the orientation may be incorrect leading to biases in the tracker. This could be addressed by adding additional components to our model and learning the model parameters from data at the expense of higher complexity.

Another issue is whether size parameters should be estimated anew at each frame. A memory of previous size estimates can be used to constrain size. However, an advantage of permitting length and width to vary is that this naturally accounts for cases of self occlusions.

A failure mode occurs if a target vehicle comes too close to another object resulting in them merging. This is sometimes a problem for parked vehicles, but rarely for moving vehicles.

## 8. Conclusion

We presented an adaptive filter designed to model LADAR hits on a wide range of vehicles. Matching involves straight-forward and fast minimization of an analytic cost function resulting in an estimate of target pose and size. Furthermore, our perturbation analysis provides a first-order uncertainty model which can be used directly by a Kalman filter or other kinematic tracker. The VDAMF naturally handles LADAR hit distribution changes due to self-occlusion at all azimuth angles, which removes the need to handle these explicitly in the optimization process and keeps the number of parameters small.

The output of the VDAMF can be interpreted both as a pose for a tracker, and also as defining a canonical coordinate system for vehicles and clutter. Aligning target objects to this coordinate system is a key step in vehicle discrimination.

There are a number of enhancements we plan for future work. Certain aspects of the VDAMF are chosen by hand, such as visible edge width. These instead could be optimized over labeled data to better match real vehicle data. In addition, a bank of filters for different vehicle types could be learned and the best filter for each target used.

## Acknowledgment

The authors greatly appreciate the help of Steve McLean in performing data collections and whose algorithms do the initial clustering of LADAR points.